%% file: arxiv.tex
\documentclass{article} 
\usepackage{bats_preprint,times}

\input{macro}

\batsfinalcopy

\begin{document}

\title{Leveraging Large Language Models \\ for Structure Learning \\in Prompted Weak Supervision}

\author{
  Jinyan Su\thanks{Equal contribution}$\ \ ^{,1}$\hspace{0.5em}Peilin Yu$^{*, \dagger, 2}$\hspace{0.5em}Jieyu Zhang$^3$\hspace{0.5em}Stephen H. Bach$^2$\\
  $^1$Cornell University $^2$Brown University $^3$University of Washington \\
  $\dagger$ Correspondence to \texttt{peilin\_yu@brown.edu} 
}

\maketitle

\input{sections/0-abstract}

\input{sections/1-intro}

\input{sections/2-background}

\input{sections/3-approach}

\input{sections/4-experiment}

\input{sections/5-ablation_and_analysis}

\input{sections/6-discussion}

\input{sections/8-acknowledgement}

\bibliography{cite}

\end{document}

%% file: macro.tex
\usepackage[T1]{fontenc}
\usepackage{xcolor}
\usepackage{tabularray} 
\usepackage{hyperref}
\usepackage{booktabs}
\usepackage{graphicx}
\usepackage{multirow}
\usepackage{algorithm}
\usepackage{algpseudocode}
\usepackage{subcaption}
\usepackage{bm}
\usepackage{marvosym}
\usepackage{float}
\usepackage{color, colortbl}
\usepackage{nicematrix}
\definecolor{LightCyan}{rgb}{0.88,1,1}
\definecolor{LighterCyan}{rgb}{0.8,1,1}
\definecolor{Gray}{gray}{0.9}

\usepackage{mathtools}
\usepackage[usestackEOL]{stackengine}
\newcommand{\splitcell}[1]{\Centerstack{#1}}
\newcolumntype{P}[1]{>{\centering\arraybackslash}p{#1}}
\newcolumntype{M}[1]{>{\centering\arraybackslash}m{#1}}
\usepackage{amssymb}
\usepackage{pifont}
\usepackage{wrapfig}
\usepackage{amsmath,amssymb,amsfonts}
\usepackage{graphicx}
\usepackage{textcomp}
\usepackage{graphicx}

\usepackage{url}
\usepackage{xspace}
\newcommand{\methodname}{Structure Refining Module\xspace}

\usepackage{amsmath,amsfonts,bm}

\def\eqref#1{equation~\ref{#1}}

\def\1{\bm{1}}

\DeclareMathAlphabet{\mathsfit}{\encodingdefault}{\sfdefault}{m}{sl}
\SetMathAlphabet{\mathsfit}{bold}{\encodingdefault}{\sfdefault}{bx}{n}



%% file: sections/0-abstract.tex
\begin{abstract}
Prompted weak supervision (PromptedWS) applies pre-trained large language models (LLMs) as the basis for labeling functions (LFs) in a weak supervision framework to obtain large labeled datasets.
We further extend the use of LLMs in the loop to address one of the key challenges in weak supervision: learning the statistical dependency structure among supervision sources.
In this work, we ask the LLM how similar are these prompted LFs.
We propose a \methodname, a simple yet effective first approach based on the similarities of the prompts by taking advantage of the intrinsic structure in the embedding space.
At the core of \methodname are Labeling Function Removal (LaRe) and Correlation Structure Generation  (CosGen).
Compared to previous methods that learn the dependencies from weak labels, our method finds the dependencies which are intrinsic to the LFs and less dependent on the data.
We show that our \methodname improves the PromptedWS pipeline by up to 12.7 points on the benchmark tasks.
We also explore the trade-offs between efficiency and performance with comprehensive ablation experiments and analysis.
Code for this project can be found in \url{https://github.com/BatsResearch/su-bigdata23-code}.
\end{abstract}

%% file: sections/1-intro.tex
Scarcity of labeled data is a crucial bottleneck for many supervised machine learning applications.
Recently, programmatic weak supervision (PWS) \citep{ratner2016data,ratner2017snorkel,zhang2022survey} emerges as a promising approach for efficiently creating large amounts of labeled data in alternative to costly manual labeling. 
PWS relies on multiple expert-designed labeling functions (LFs) from rules or heuristics to act as weak supervision sources and vote for a label. 
In practice, given unlabeled data, a PWS system infers probabilistic labels by resolving the agreements and disagreements among the noisy LFs through label modeling.
Recently, a new paradigm called prompted weak supervision (PromptedWS) \citep{smith2022language, yu:acl23} has been proposed to integrate pre-trained large language models (LLMs) with PWS.
In PromptedWS, program-based LFs written in code are replaced with prompted LFs: natural language prompts in which their respective labeling decisions are made through LLM inferences.
However, the predictions from LLMs can be strongly correlated due to model biases, leading to inferior labeling performance. 
Accurately estimating and managing the dependency structure among prompted LFs presents a challenging problem.
\input{figures/fig-6-framework}
Programmatic weak supervision has been successful for various applications in information extraction, medical imaging and sequence tagging~\citep{bach2019snorkel,suri2020leveraging,re2019overton,kuang2022firebolt,safranchik2020weakly}.
Recently, there have been a growing interest in integrating powerful pretrained LLMs into weak supervision workflows \citep{smith2022language,arora2022ask,zhang2022prboost}.
prompted weak supervision (PromptedWS) \citep{smith2022language} presents a unique opportunity to replace rigid code-based LFs with more flexible natural language prompts.
While this approach offers a much more flexible paradigm for weak supervision sources, prompted LFs may be susceptible to model bias and strong correlations due to their reliance on the outputs of LLMs.
Therefore, it is crucial to identify and manage the dependency structures among the prompted LFs to ensure accurate label estimation.

While structure learning in the supervised learning settings has been extensively studied \citep{meinshausen2006high,zhu2010grafting}, learning the structure in programmatic weak supervision presents significant challenges due to the absence of the true labels \citep{bach2017learning}. 
The difficulties are further exacerbated in a prompted weak supervision setup, where individual supervision sources are expressed as natural language prompts for LLMs, the sole shared knowledge source.
To handle the correlations of labeling functions (LFs), early work relies on user-specified dependency structures \citep{ratner2016data}. 
Previous studies have also explored how to automatically discover the structure with only unlabeled data \citep{bach2017learning,varma2017inferring}. 
However, existing methods are not designed to handle correlations in Prompted weak supervision setup.
In prompted weak supervision, weak labels are solely based on querying prompted LFs via the Large Language Models and the model responses alone may not fully represent the correlations among the prompted LFs.

In this paper, we propose a \methodname, a simple yet effective structure learning method that plugs into existing PromptedWS workflows (Figure~\ref{WS_framework}).
The key idea is to guide and accelerate the structure learning process by using the LLM's internal representation of prompted LFs.
We find that the similarity of the representations of the prompted LFs is predictive of harmful correlations.
Further, they can be computed much more quickly than executing the prompted LFs on many examples.
Our approach contains two core components:  ~\textit{\textbf{La}beling function \textbf{Re}moval} (LaRe) and ~\textit{\textbf{Co}rrelation \textbf{S}tructure \textbf{Gen}eration} (CosGen).
LaRe is used for automatically detecting ~redundant prompted LFs to reduce both LLM query cost and avoid severe biases.
CosGen is specifically designed to efficiently learn the dependency structures among prompted LFs, utilizing only the vector embeddings of the prompted LFs themselves, without relying on any labels or unlabeled data.
In practice, LaRe will first be applied on the given set of prompted LFs to detect and remove any unnecessary prompted LFs.
Then CosGen will produce the dependency structure for label modeling.
Our approach allows the end users to quickly and accurately discover the correlations while significantly reducing the computational cost for both structure learning and label modeling.

To summarize, our main contributions are three-fold:
\begin{enumerate}
\item We propose a novel method for efficient structure learning in prompted weak supervision by leveraging similarity in the embedding space of prompted labeling functions. 
\item We demonstrate the effectiveness of our proposed method by showing that it outperforms the prompted weak supervision baseline by an average of 5.9 points while saving significant computation.
\item We conduct comprehensive ablation and analysis experiments. We validate the core assumptions and analyze the contribution of each component of our method.
\end{enumerate}

%% file: figures/fig-6-framework.tex
\begin{figure*}[ht!]
    \centering
    \includegraphics[width=1.0\textwidth]{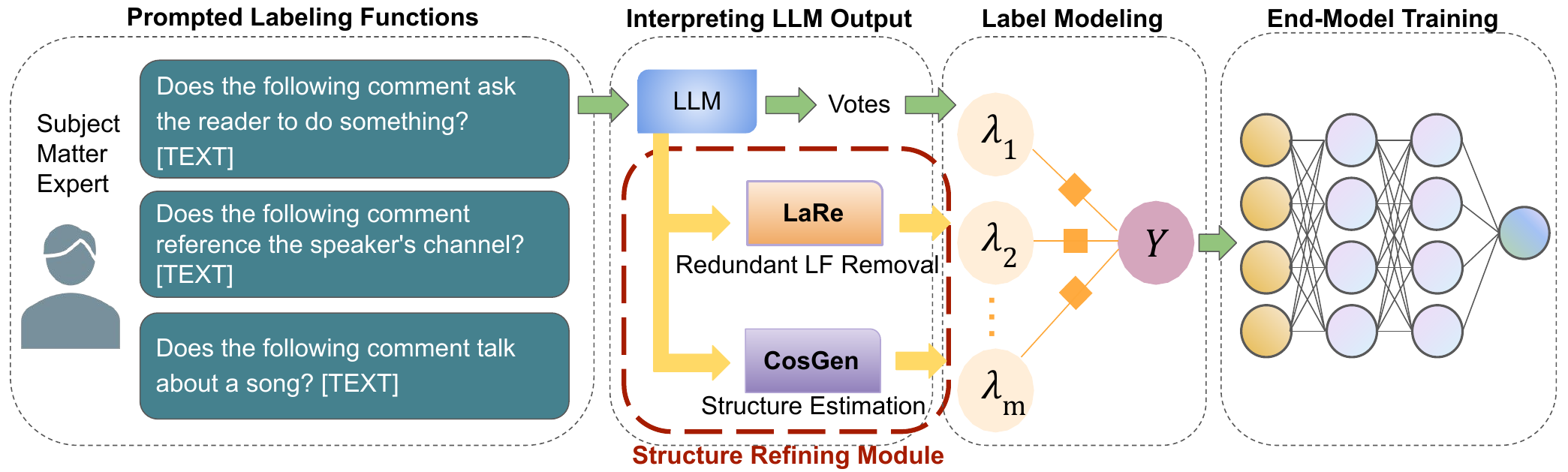}
    \caption{Prompted weak supervision workflow showing the \methodname\ as a plugin.}
    \label{WS_framework}
\end{figure*}

%% file: sections/2-background.tex
\section{Background}\label{sec3}

\subsection{Problem Setup}
\noindent
\textbf{Prompted Weak Supervision:}
In a prompted weak supervision (PromptedWS) setup, given a dataset ~$D=\{x_1,x_2,\cdots,x_n\}$ of classification task with $n$ unlabeled data points and each $x_i (i \in [n])$ with an unobserved true label $y_i\in\mathcal{Y}$, the goal is to infer the true label using weak supervision.
We are also given a small set of heuristics created by subject matter experts (SMEs) manually for labeling the unlabeled data.
For example, in the case of labeling spam comments, the SME may find that many spam examples contains call to action, such as asking readers to subscribe to a channel or asking the readers to buy something.
Then she may create a heuristic such as ``Does the following comment ask the reader to do something?".
In PromptedWS, these heuristics are formatted as prompts, which are then fed into an instruction-tuned LLM such as T0++~\citep{sanh2021multitask} to get the answer. 
After that,  we map the answer to a label or abstain according to the SMEs' heuristics.
Formally, by observing the features in the development set, the SME could design a set of heuristics
$\mathcal{H} = \{h_1,\cdots,h_m\}$ as well as a label map $\mathcal{M}: \mathcal{S}\rightarrow \mathcal{Y}\cup \{\emptyset\}$. The heuristics are encapsulated by a prompt template. 
For example, in the case of website comment, the prompt template would be
\begin{center}  
\small{\texttt{Does the following comment ask the reader to do something? [TEXT]}}
\end{center}
where \texttt{[TEXT]} is a placeholder for the text of each comment. 
By feeding this template to the pre-trained LLM $\mathcal{A}$, it returns a text string: $s = \mathcal{A}(h,x) \in \mathcal{S}$, where  $\mathcal{S}$ is the set of all possible strings that could be returned from the LLM. 
Then the label map $\mathcal{M}$ maps the returned answer $s$ to one of the class labels within the label space $\lambda_i \in \mathcal{Y}$ to vote for a label or maps $s$ to a special symbol $\emptyset$, indicating abstaining and not voting for any label. 
In the case of the above example prompt, the label map would map positive responses such as \texttt{YES} and \texttt{True} to the SPAM label and all the other responses to abstentions.

Combining the above prompt template and label map together, we get a set of \textit{Prompted labeling functions}: $S_{\text{LF}}=\{\mathcal{M}(\mathcal{A}(h_1,\cdot)), \cdots,\mathcal{M}(\mathcal{A}(h_m,\cdot))\}$. 
Applying the above $m$ LFs to the unlabeled dataset $D=\{x_1,\cdots,x_n\}$ would create a $n\times m$ label matrix $L$. 
The rest of the workflow remains the same as a standard weak supervision framework: a \textit{label model} is used to aggregates the votes in $L$ to produce probabilistic estimates of the true label.
While many label modeling methods assume conditional independence among the LFs given the latent label~\citep{ratner2016data,ratner2017snorkel}, recent label models may consider dependency structures provided by the user~\citep{fu2020fast}.
Finally an appropriate \textit{end model}, such as a deep neural network, is trained by minimizing the expected empirical risk with respect to the probabilistic estimates of the true labels.

\subsection{Related Work}

\noindent
\textbf{Programmatic Weak Supervision:} Manually labeling training data is prohibitively expensive and time-consuming. 
A common alternative is to use weak supervision sources.
Estimating the accuracy of multiple sources is well studied in many areas such as crowd sourcing\citep{dalvi2013aggregating,joglekar2015comprehensive}, boosting \citep{schapire2013boosting,balsubramani2015scalable,zhang2016spectral}, and co-training \citep{blum1998combining}. 
In this paper, we focus on programmatic weak supervision \citep{ratner2016data,ratner2017snorkel,zhang2022survey}, which estimates the accuracy of weak sources without \textit{any} labeled data. 
In programmatic weak supervision, users encode various weak supervision sources such as user-written heuristics \citep{ratner2017snorkel,meng2018weakly,awasthi2020learning}, knowledge bases \citep{liang2020bond,hoffmann2011knowledge}, pre-trained models \citep{bach2019snorkel,zhang2021creating,yu2022nplm,smith2022language}, and third-party tools \citep{lison2020named} into labeling functions (LF), which can then give votes on the true labels of the unlabeled data.

\noindent
\textbf{Integrating LLMs into Weak Supervision:}  
Prompting is a powerful technique for many few-shot and zero-shot applications with large language models \citep{liu:arxiv21}.
Prompting has been used to create and modify datasets in many ways \citep{schick:emnlp21,ye:arxiv22,chia:acl22,wu:acl22,bonifacio:arxiv22,lang:arxiv22,elazar:tacl21,zhang:acl22}.
Recent studies explored integrating prompting and weak supervision \citep{smith2022language}.
Prompting also offers a unique opportunity to relax rigid code-based programmatic weak supervision sources.
In this work, we aim to make the integration of LLMs into weak supervision more effective by mitigating the downside of hazardous correlations among the prompted LFs.

\noindent
\textbf{Structure Learning in Weak Supervision:} It can be highly beneficial to capture the statistical dependencies among LFs during label modeling since model misspecification can lead to incorrect estimates of the true labels \citep{cachay2021dependency}. 
Multiple methods for learning such dependencies have been proposed \citep{bach2017learning,varma2017inferring,varma2019learning}.
Bach et al.\ \citep{bach2017learning} and Varma et al.\ \citep{varma2019learning} learn dependency structures from the votes of the LFs.
Varma et al.\ \citep{varma2017inferring} infer the structure through static analysis of the code specifying the LFs, with additional assumptions that the LFs operate over domain-specific primitives. 
With the dependency structure in hand, many methods can then incorporate this information into the label modelling process for improved label quality, either through factor functions \citep{ratner2016data,shin2021universalizing} or graph structures \citep{ratner2019training,fu2020fast,varma2019multi}.
The structure obtained by our \methodname\ can be incorporated into any such method.

%% file: sections/3-approach.tex
\section{Approach}
\label{sec6}

In this section, we describe \methodname for estimating the dependency structure among prompted LFs in the PromptedWS setting.
We explore the potential of Large Language Models beyond its current usage in PromptedWS by leveraging its powerful representation capabilities.
The core idea is to extract embeddings in addition to the inference output.
Our method assumes that LFs with high similarity in the latent representation space are more likely to be correlated.
Specifically, for each prompted LFs, we first extract vector embeddings from the last hidden layer of the LLMs. 
Then we can compute a pairwise similarity matrix $M$ based on the extracted embeddings for all prompted LFs.
Specifically, we encapsulated our method into one plug-in for PromptedWS.
Our method contains two sequential components: ~\textit{\textbf{La}beling function \textbf{Re}moval} (LaRe) and ~\textit{\textbf{Co}rrelation \textbf{S}tructure \textbf{Gen}eration} (CosGen).
LaRe is designed to remove heavily correlated LFs while balancing performance and efficiency.
It can be especially helpful when the total number of LFs is very large.
In addition to LaRe, we introduce CosGen to estimate the dependency structures for the prompted LFs.
We include the pseudocode in \ref{alg1}.
Next, we describe the LaRe and CosGen in detail.

\input{figures/main-algo}

{\renewcommand{\arraystretch}{1.2}
\begin{table*}[h]
    \centering
    \caption{Summary statistics for text classification dataset used in our experiments. P(positive) is the class balance of the positive label calculated by the relative frequency of all gold labeled splits with standard error in the parentheses. Note that since we focus on PromptedWS, all the labeling functions we refer to here are prompted LFs as defined in \citet{smith2022language}.}
    \resizebox{0.98\linewidth}{!}{
  		\begin{tabular}{|c|c|c|c|c|ccc|}
			\hline
\textbf{Name}	&\textbf{Task}		& \textbf{Class}& \textbf{P(positive)}&\textbf{\#LFs}  &\textbf{\#Train}&\textbf{\#Validation}& \textbf{\#Test}\\
\hline
		YouTube&Spam Detection& HAM,SPAM&0.488(0.02)&10& 1586&120&250\\
		SMS&Spam Detection&HAM,SPAM&0.132($<0.01$)&73& 4571&500&500\\
			Spouse&Relation extraction&NOT SPOUSE, SPOUSE&0.074($<0.01$)&11& 22254&2811&2701\\
			
			\hline
		\end{tabular}}
  
    \label{tab:datasets}
\end{table*}
}

\input{figures/tables/tab-1-main-results}

\noindent \textbf{LaRe:}
The goal of LaRe is to remove redundant prompted LFs in order to reduce computation need while maintaining labeling accuracy. 
To begin, the user may decide how many LFs should be removed by examining the similarity matrix among the prompted LFs set $S_{\text{LF}}$.
Higher concentrations in the similarity matrix indicate higher correlations or possibility of redundant LFs to be removed.
In addition, more LFs could lead to more redundant LFs to be removed when the labeling function set $S_{\text{LF}}$ is large.
After deciding how many labeling functions to be removed, potentially redundant LFs are then identified, removed based on their similarity matrix $M$.
First we will rank the pairwise similarities for each LF pair.
Then at each step, we find the LF pair that has the largest similarity and remove the LF with a larger index to keep our algorithm deterministic.
Removing the redundant LF results in a new set of LFs $S_{\text{LF}}^{'}$, where $|S_{\text{LF}}^{'}| = m-m_r$ and $m_r$ is the total number of LFs we removed. 
We include detailed investigations into how to effectively remove redundant LFs in Section \ref{sec5}. \\

\noindent \textbf{CosGen:} 
CosGen aim to estimate the dependency among LFs based on the correlations indicated in the similarity matrix $M^{'}$, which contains only the correlation information of the refined LF set $S^{'}_{\text{LF}}$. 
One key intuition is to construct dependency for highly correlated LFs indicated by  $M^{'}$.
However, the structure directly estimated  from this method may be non-identifiable and can not be used for label modeling. 
In order to guarantee the structure to be identifiable, we first find two LFs that have the smallest correlation and assume these two to be mutually independent while at the same time, also to be independent with all the other LFs. 
Then we find the rest dependencies by assigning edges to LFs with high similarity.
The main hyperparameter for CosGen is the number of edges ($m_e$) we want to have in our graph and it decides how sparse the graph is. 
In practice, we can also replace this parameter to a percentage threshold on the total number of possible edges.
Note that $m_e\leq \frac{(m-2)\cdot (m-3)}{2}$. 
The dependency structure inferred from our similarity matrix can serve as the input to any probabilistic label model that supports source dependency modeling. 

%% file: figures/main-algo.tex
\begin{algorithm}
	\caption{Pseudocode for \methodname}
	\begin{algorithmic}[1]
		\State {\bfseries Input:} Similarity matrix $M$, original Prompted LF set $S_{\text{LF}}$, number of LF to be removed: $m_r$, number of edges in the graph: $m_e$.
 \State $S_{r}=\emptyset$, $\mathcal{E}=\emptyset$\\
 $//$ Labeling function removal (\textbf{LaRe})
 \For{t  $\in [m_r] $}
 \State $(i, j)=\arg\max M_{i,j}$
 \State $k=\max\{i,j\}$
 \State $S_r\leftarrow S_r\cup \{k\}$
 \State $ M_{k,:}\leftarrow 0, M_{:,k}\leftarrow 0$
\EndFor
\State $S^{'}_{\text{LF}}\leftarrow S_{\text{LF}}/S_r$
\\
	$//$ Correlation structure generation (\textbf{CosGen})
 \State Let $M^{'}$ be the similarity matrix of LFs in $S^{'}_{\text{LF}}$
 \State $(i,j) = \arg\min M^{'}_{i,j}$
 \State $M^{'}_{i,:}\leftarrow 0$, $M^{'}_{:,i}\leftarrow 0$, $M^{'}_{j,:}\leftarrow 0$, $M^{'}_{:,j}\leftarrow 0$
 \For{$t \in [m_e]$ }
 \State $(i, j)=\arg\max M^{'}_{i,j}$
 \State $\mathcal{E} = \mathcal{E}\cup\{(i,j)\}$
 \State $ M^{'}_{i,j}\leftarrow 0, M^{'}_{j,i}\leftarrow 0$
\EndFor
\\
	 \Return Refined labeling function set $S^{'}_{\text{LF}}$, edges in the graph $\mathcal{E} $
	\end{algorithmic}
	\label{alg1}
\end{algorithm}

%% file: figures/tables/tab-1-main-results.tex
\begin{table*}
    \centering
    \caption{Main results from WRENCH benchmark, PromptedWS, PromptedWS with data based weak supervision structure learning method (WSSL) and PromptedWS with \methodname (ours). For all the experiments, we run 5 random seeds and reports the mean and standard error.  }
    \vspace{0em}
  		\begin{tabular}{|c|cc|c |c |}
			\hline
& & \textbf{YouTube(Acc)} & \textbf{SMS(F1)} &\textbf{Spouse(F1)} \\
\hline
\splitcell{WRENCH  \citet{zhang2021wrench}}&& $94.6\pm0.5$& $93.5\pm0.7$&$25.5\pm1.8$\\
\splitcell{PromptedWS  \citet{smith2022language}}& &$94.8\pm1.2$& $90.0\pm4.1$&$52.1\pm1.3$\\
\splitcell{PromptedWS + WSSL  \citet{varma2019learning}}&& $93.0\pm1.3$ &\textbf{94.5}$\pm$1.7 & $63.9\pm0.9$ \\
\multirow{1}{*}{\splitcell{PromptedWS + \\ \ \methodname}}& & \textbf{95.6}$\pm$0.4 &$94.2\pm1.4$  
   &\textbf{64.8}$\pm$0.2\\
			\hline
		\end{tabular}
    
    \label{ta7}
\end{table*}

%% file: sections/4-experiment.tex
\section{Experimental Results} \label{sec7}

\noindent
\textbf{Dataset}
In our experiments, we evaluated on three datasets: YouTube \citep{alberto2015tubespam}, SMS \citep{gomez2006content}, and Spouse \citep{corney2016million}.
(See the summary of the datasets in Table \ref{tab:datasets}.) 
We follow the prompted labeling functions from \citep{smith2022language}, which are translated from labeling functions from WRENCH \citep{zhang2021wrench}, a standard weak supervision benchmark.
We include all natural language prompts for our experiments in \ref{tab:LFs}

\noindent
\textbf{Baselines} We empirically evaluate the our method against three baselines:
(1) the standard weak supervision with labeling functions in WRENCH benchmark \citep{zhang2021wrench}. 
(2) PromptedWS \citep{smith2022language} (without our refining module)
(3) PromptedWS with weak supervision structure learning (WSSL) \citep{varma2019learning} by passing the structure learned from WSSL to the label model in the PromptedWS)
We report default metrics specified in the WRENCH benchmark \citep{zhang2021wrench} for direct comparison:
for YouTube dataset, we report accuracy and for SMS and spouse dataset, we report F1.
Our performance metrics are reported as the mean and standard error of 5 independent runs using different random seeds.

\noindent
\textbf{Setup} We follow the conventional workflow of programmatic weak supervision and compare 4 approaches relevant to programmatic weak supervision and structure learning. 
For each dataset in our analysis, we assume the training splits are unlabeled, the LFs (either prompted LFs in PromptedWS or code-based LFs from WRENCH benchmark) are applied to the unlabeled training splits to generate weak labels. 
Throughout our experiments, we use T0++ \citep{sanh2021multitask}, a 11 billion parameter instruction-tuned large language model based on T5 \citep{raffel2020exploring} architecture.
T0++ is trained using large dataset of supervised tasks transformed into instruction prompted training data and can achieve competitive zero shot classification performance. 
T0++ model requires 42 GB of GPU to efficiently run locally without parameter offloading. We use 2 NVIDIA A100 GPUs for inference.
In all of our experiments, unless otherwise specified, we use Flyingsquid \citep{fu2020fast} as our label model to combine and denoise the labelers' votes into probabilistic labels and incorporate infered LFs dependency structures.
The resulting probabilistic labels are then used to train an end classification model based on RoBERTa \citep{liu2019roberta}.

\noindent
\textbf{Results} Table \ref{ta7} outlines the performance of the 3 baselines (WRENCH, PromptedWS, PromptedWS+WSSL) and as well as our method (PromptedWS+\methodname). 
The hyperparameters for LaRe, CosGen and the end model are chosen by performance on the validation dataset.
For Spouse dataset, our method improves 39.3 F1 points compared to WRENCH benchmark, and improved 12.7 F1 score compared to the PromptedWS.
For Youtube, \methodname increase 1.0 and 0.8 Acc points in comparison to  WRENCH and PromptedWS respectively while surpassing PromptedWS+WSSL by 2.6 Acc points.
For SMS dataset, \methodname also provides performance advantage ovber WRENCH and PromptedWS by 0.7 and 4.2 F1 points. 
On average, our method improves 13.7 points compared to WRENCH benchmark, 5.9 points compared to PromptedWS and 1.1 points compared to PromptedWS+WSSL over the three datasets. 
Besides potential performance advantage compared to WSSL, \methodname can be more efficient in both time and computation.
We provide further analysis on performance efficiency of \methodname in Section \ref{Efficiency}.

Overall, our experimental results demonstrate that \methodname can significantly improve the performance over the PromptedWS which doesn't take into account the correlations among prompted LFs. 
In addition, the PromptedWS in complementary with our method can achieve superior performance, outperforming the WRENCH benchmark in all the three datasets used in our experiment. 
Most notably in Spouse dataset, we improved 39.3 F1 points over WRENCH benchmark. 
Our method outperforms or achieves comparable result to the state-of-the-art structure learning method WSSL \citep{varma2019learning}, indicating that the correlations found using our refining model are more effective or comparable to learning the structure from data. 

\input{figures/tables/tab-8-lfs}

%% file: figures/tables/tab-8-lfs.tex
{\renewcommand{\arraystretch}{1.2}
\begin{table*}[h]
    \centering
    \caption{Prompted Labeling Functions for YouTube and Spouse datasets. The YouTube dataset has class labels HAM and SPAM, and the dataset has class labels NOT SPOUSE and SPOUSE. The label map
transforms the answer “yes” to the value denoted by the label column(either HAM or SPAM for YouTube dataset and either NOT SPOUSE or SPOUSE for spouse dataset) and
we consider other answers as abstention.
for the SMS dataset, The template asks whether there are certain keywords in the text and the label map transforms the answer ``yes" to the value denoted by the label column (either HAM or SPAM) and we consider other answers as abstention.
prompted Labeling Functions are from ~\citet{smith2022language}.}
    \vspace{0em}
    \resizebox{1.0\linewidth}{!}{
  		\begin{tabular}{|c|l|c|}
			\hline
\textbf{Dataset}&\textbf{Template}		& \textbf{Label}\\ \hline
	\multirow{10}{*}{\splitcell{\textbf{YouTube}}} 
&	Does the following comment talk about a song?\textbackslash n\textbackslash n[TEXT] &HAM\\
&	Is the following comment fewer than 5 words?\textbackslash n\textbackslash n [TEXT]& HAM\\
&Does the following comment mention a person’s name?\textbackslash n\textbackslash n [TEXT]                            &HAM \\
&	Does the following comment express a very strong sentiment?\textbackslash n\textbackslash n [TEXT]& HAM\\
&Does the following comment express a subjective opinion?\textbackslash n\textbackslash n [TEXT]                &HAM\\
&Does the following comment reference the speaker’s channel or video? \textbackslash n\textbackslash n [TEXT]    & SPAM\\ 
&Does the following comment ask you to subscribe to a channel?\textbackslash n\textbackslash n [TEXT]            &SPAM \\
&Does the following comment have a URL?\textbackslash n \textbackslash n[TEXT]                                       & SPAM \\
&Does the following comment ask the reader to do something?\textbackslash n\textbackslash n [TEXT]                        &SPAM \\
&Does the following comment contain the words "check out"? \textbackslash n\textbackslash n [TEXT]                    &SPAM \\
			 \cline{2-2}
   	\hline
				\multirow{10}{*}{\splitcell{\textbf{Spouse}}} 
						&Context: [TEXT]\textbackslash n \textbackslash n Are [PERSON1] and [PERSON2] family members? &NOT SPOUSE \\
			&Context: [TEXT]\textbackslash n \textbackslash n Is [PERSON1] said to be a family member?& NOT SPOUSE \\
			&Context: [TEXT]\textbackslash n \textbackslash n Is [PERSON2] said to be a family member?& NOT SPOUSE \\
			&Context: [TEXT]\textbackslash n \textbackslash n Are [PERSON1] and [PERSON2] dating? &NOT SPOUSE\\
		&Context: [TEXT]\textbackslash n \textbackslash n Are [PERSON1] and [PERSON2] co-workers?& NOT SPOUSE\\
		&Context: [TEXT]\textbackslash n \textbackslash n Is there any mention of "spouse" between the entities [PERSON1] and [PERSON2]? &SPOUSE \\
			&Context: [TEXT]\textbackslash n \textbackslash n Is there any mention of "spouse" before the entity [PERSON1]? &SPOUSE \\
		&Context: [TEXT]\textbackslash n \textbackslash n Is there any mention of "spouse" before the entity [PERSON2]?& SPOUSE\\
		&Context: [TEXT]\textbackslash n \textbackslash n Do [PERSON1] and [PERSON2] have the same last name? &SPOUSE \\
		&Context: [TEXT]\textbackslash n \textbackslash n Did [PERSON1] and [PERSON2] get married? &SPOUSE \\
		& Context: [TEXT]\textbackslash n \textbackslash n Are [PERSON1] and [PERSON2] married?& SPOUSE\\
            \hline
		\end{tabular}}

\renewcommand{\arraystretch}{1.2}
  		\begin{tabular}{c}
			\\
		\end{tabular}
  
    \resizebox{1.0\linewidth}{!}{
  		\begin{tabular}{|l|l|l|}
			\hline
\textbf{Template for SMS}  &\textbf{Does the following text message contain the words "[KEYWORDS]"?\textbackslash n \textbackslash n [TEXT]}& \textbf{Label}\\ \hline
\multirow{10}{*}{[KEYWORDS]} 	& \multicolumn{1}{|p{12cm}|}	
 {??1.50, ??500, ??5000, call for offer, cash prize, chat date, chat to, childporn, credits, dating call, direct, expires now, fantasies call, free phones, free price, free ringtones, free sex, free tone, guaranteed free, guaranteed gift, hard live girl, important lucky, inviting friends, latest, latest offer, message call, new mobiles, no extra, password, please call, sms reply, unlimited calls, urgent award guaranteed, urgent prize, voucher claim, welcome reply, win shopping, winner reward, won call, won cash, won cash prize, won claim}
 & SPAM\\
 \cline{2-3}
 & \multicolumn{1}{|p{12cm}|}
{I, I can did, I it, I miss, I used to, adventuring, amrita, can’t talk, did u got, do you, fb, goodo, hee hee, i’ll, jus, link, maggi, mine, my kids, noisy, praying, shit, should I, thanks, that’s fine, thats nice, u how 2, we will, where are, wtf, your I}

 &HAM\\
			\hline
		\end{tabular}}
    
    \label{tab:LFs}
\end{table*}

%% file: sections/5-ablation_and_analysis.tex
\section{Ablation and Analysis}
\label{sec5}

In this section, we provide further analysis on our core assumption that high similarities in LFs embedding similarities suggest correlation.
We also exam the contribution of LFs removal and correlation structure generation from similarities from both performance and efficiency perspectives.
We conduct experiments to evaluate the efficiency trade-offs for label modeling in our settings and lastly we provide intuitions on how to set the hyperparameters for \methodname. 

\input{sections/5-1-sim2corr}

\input{sections/5-2-LFs_removal}

\input{sections/5-3-struct_gen}

\input{sections/5-4-augmentated_lfs}

\input{sections/5-5-effect_lm}

\input{sections/5-6-hyperparam_guide}

%% file: sections/5-1-sim2corr.tex
\subsection{Similarities reveal correlations}
\input{figures/tables/tab-5-toy-experiment-s2c}
\input{figures/fig-3-sim-viz}
In this subsection, we aim to show that LFs similarities in the embedding spaces are valid indicators for their correlations.
Previous work \citep{bach2017learning,varma2019learning} characterizes LFs correlation structures based on their output, which can be sensitive to the data and can result in spurious correlations. 
We aim to learn more expressive correlations that are invariant to data.
To achieve this, we focus on finding correlations intrinsic to the LFs.
Here we consider each LF as a task and capture their correlations through their respective task embeddings. 
We use T0++ \citep{sanh2021multitask} to embed each LFs with last layer embeddings to compute cosine similarities matrix.
The heatmap visualization of the similarity matrix for the YouTube dataset is shown in Figure \ref{fig:youtube_similarity_df}.
As demonstrated by the figure, the similarity information has significant overlaps with the correlated errors among LFs, which can be used to remove the most correlated LFs and still achieve good performance.
In addition, we construct a toy experiment to further exam our assumption.
This experiment is based on the intuition that if two labeling functions (LFs) are correlated, removing one should not significantly harm the model's performance.
Removing the most correlated LF leads to more accurate weights assigned to each LF, resulting in a more accurate end model. 
The results from our toy experiment align with the assumption. 
As shown in Table \ref{ta4}, after removing one of the most correlated LFs, the result either improved or slightly dropped below the baseline. 
This investigation further suggests that the similarity information in the latent embedding space can reveal information about the underlying correlations of LFs.

%% file: figures/tables/tab-5-toy-experiment-s2c.tex
{\renewcommand{\arraystretch}{1.2}
\begin{table}[h]
    \centering
    \caption{Toy experiment of removing one of the most correlated LFs. The first row is the result of PromptedWS without moving any LFs, and the second and third rows are the result of PromptedWS after removing one LF from the most correlated LFs pairs.}
    \resizebox{0.75\linewidth}{!}{
  		\begin{tabular}{|l| c|c |c|}
			\hline
   
&  \textbf{YouTube} & \textbf{SMS} & \textbf{Spouse} \\
\hline
PromptedWS& 94.8(1.2)& 90.0(4.1)&52.1(1.3)\\
\hline
\multirow{2}{*}{\splitcell{Removing most correlated LF}}& 94.7$\pm$0.7\  &93.0$\pm$0.8$\textcolor{red}{\uparrow}$ & 53.0$\pm$1.2$\textcolor{red}{\uparrow}$\\
& 95.3(0.3)$\textcolor{red}{\uparrow}$& 91.4(2.2)$\textcolor{red}{\uparrow}$&52.3(1.8)$\textcolor{red}{\uparrow}$\\
			\hline
		\end{tabular}}
    \label{ta4}
\end{table}
}

%% file: figures/fig-3-sim-viz.tex
\begin{figure}[ht]
    \centering
    \includegraphics[width=\linewidth]{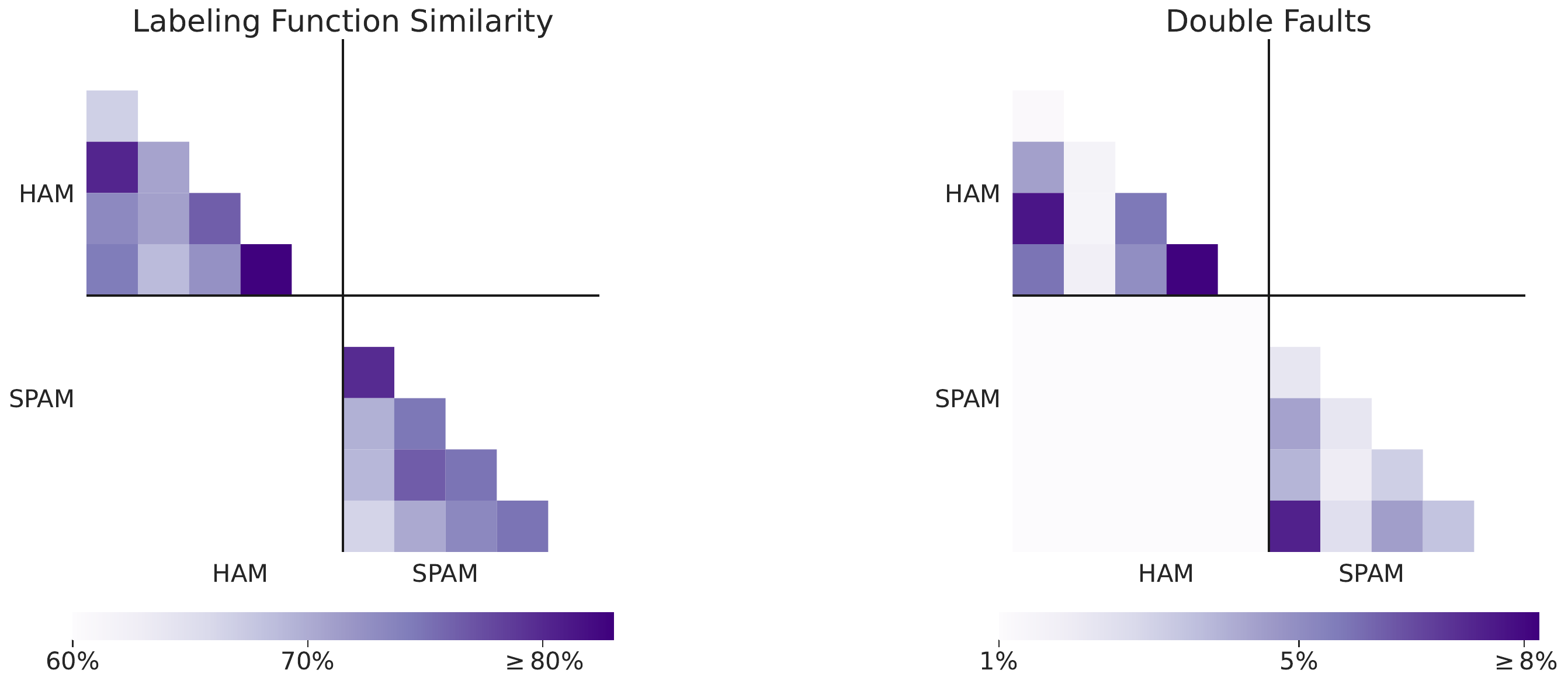}
    \caption{Visualization of the similarity matrix for labeling functions in the YouTube dataset compared with double faults, i.e., examples on which both labeling functions make a mistake.}
    \label{fig:youtube_similarity_df}
\end{figure}

%% file: sections/5-2-LFs_removal.tex
\subsection{Role of Labeling Function Removal}
In this subsection, we exam the contribution of Labeling function Removal (LaRe) in our \methodname.
The core idea for LaRe is that when LFs are highly correlated and redundant, removing some can improve performance and efficiency.
To extensively understand effect of LFs removal, we can set $m_e = 0$ for our \methodname to only use LaRe. 
We examined the impact of removing LFs based on correlation by conducting experiments where we removed 10\%, 30\%, 50\%, and 70\% of labeling functions. 
\input{figures/tables/tab-2-ablation-lf-percent-removal}

As shown in \ref{pic3}, using LaRe alone could improve the performance over vanilla PropmtedWS.
Even though the original labeling functions used in PromptedWS are carefully selected to avoid correlation and facilitate weak supervision, remove 10\% of the labeling functions (on YouTube and SMS) and remove 30\% of the labeling functions (on Spouse) can still improve the performance, further suggesting that there still could be some redundancies with manually selected LFs.
Note that our experiments are relatively coarse by setting the removal rate to be 10\%, 30\%, 50\%, 70\% for all the dataset because the total number of labeling functions varies from 10 to 146 for the six group experiments, so the effects of ``removing 70\%" could be different on different dataset based on the total number of LFs as well as the redundancies of the total LFs set. 
In practice, the number or portion of removal should be more fine-grained and tailored to the total number of labeling functions to achieve a better performance. 
We expect to remove redundant labeling functions without tuning hyperparameters on label model and end model, so we apply the same hyperparameters set used for PromptedWS when doing our experiments on removal. 
In practice, as in our experiments, we use the validation split from WRENCH to evaluate, compare and choose the appropriate removal threshold.
Overall, the experiments show that the LaRe in our \methodname can help improving the performance.
These experiments demonstrate that the \textit{removing} part in our refining module could help to improve the performance even in the worst case.  
Note that whether the performance improve or drop depends on the trade-off between the performance gain by handling the dependencies and the performance drop due to the reduced information from removing labeling functions, so the performance plot over removal isn't perfectly concave. 
Also, we should not only care about the removal threshold that achieves the best performance, but all the removal thresholds where the performance are better or equivalent to the PromptedWS (namely, the all the points above or near the dashed line in Figure \ref{pic3}), since these are points indicating an improved efficiency without negatively affect performance.
\input{figures/fig-1-ablation-performance-removal-rate}

\noindent
\textbf{Performance and efficiency: trade-off or win-win?}
Inference with Large Language Models (LLMs) can be computationally expensive and time-consuming.
One benefit of removing redundant LFs is efficiency: when the total number of labeling functions is reduced, we are feeding less tokens to the LLMs and therefore save more computations and time.  
When weighing performance and efficiency, it's important to consider the feasibility of removing LFs in terms of the effect on performance. 
However, as our experiments show, it may not always be a trade-off; Sometimes it could be a win-win scenario with \methodname. 
As shown in Figure \ref{pic3}, for YouTube dataset, removing 10\% of the labeling functions improves the performance while saving more computations. 
For SMS, removing 10\% or 30\% both improve performance.  
In the most extreme cases of removing 70\% of the labeling functions, though the performance decreases 1.3 points, it reduces 51 labeling functions, and saved in total of 233121 prompts.
For the Spouse dataset, removing 30\% of LFs presents a win-win scenario, both improving performance and saving computation. 
We include the detailed statistics in table \ref{Main result1}.

%% file: figures/tables/tab-2-ablation-lf-percent-removal.tex
{\renewcommand{\arraystretch}{1.2}
\begin{table*}[t!]
    \centering
     \caption{We documented the number of labeling functions removed as well as accuracy or F1 metric for each experiment. We report the mean and standard error with 5 random runs and the best results are indicated in bold. We highlight results that outperform PromptedWS with arrows. We also estimate the total number of token saved when running with Train/Test/Validation splits.}
    \resizebox{\linewidth}{!}{
  		\begin{tabular}{|l|l|ccc|ccc|}
			\hline
&&&&&\multicolumn{3}{c|}{\textbf{Estimated \# saved tokens}}\\
\hline
&&\textbf{Performance(Acc/F1)}&\textbf{\#(LF removed)}& \textbf{\# (prompts saved)}& \textbf{Train} & \textbf{Test} & \textbf{Validation}\\
\hline
\multirow{5}{*}{Youtube} & PromptedWS & 94.8 $\pm$ 1.2&0 & 0&  0 &0&0\\
& removing 10\%&  \textbf{95.3 }$\pm$ \textbf{0.3} $\textcolor{red}{\uparrow}$ &1 & 1586&70900 & 11369&5186\\
& removing 30\%& 90.5$\pm$1.7  &  3 & 4758&212700 &  34107 &   15558       \\
& removing 50\%& 87.5$\pm$1.9&5 &7930& 354500 & 56845  & 25930    \\
&removing 70\%& 83.1$\pm 0.5$& 7&11102& 496300 &  79583   &  36302      \\
\hline
\multirow{5}{*}{SMS} & PromptedWS& 90.0 $\pm$ 4.1&0 & 0&  0 &0&0\\
& removing 10\% & \textbf{94.2}$\pm$ \textbf{1.4} $\textcolor{red}{\uparrow}$ &7 & 31997 &1343069& 144984& 147903\\
& removing 30\%& 93.5$\pm$ 0.8 $\textcolor{red}{\uparrow}$ &  22  & 100562 &4221074& 455664 &
464838 \\
& removing 50\%&   88.6$\pm$3.4&36  &164556 &6907212& 745632&760644 \\
&removing 70\%& 88.7$\pm 3.1$ &51&233121&9785217& 1056312&  1077579\\
\hline
\multirow{5}{*}{Spouse} & PromptedWS& 52.1 $\pm$ 1.3 &0 & 0&  0 &0&0\\
& removing 10\% & 53.0$\pm$ 1.2 $\textcolor{red}{\uparrow}$ &1 & 22254 &
1980962& 234659 & 249159 \\
& removing 30\%& \textbf{58.5}$\pm$ \textbf{1.3} $\textcolor{red}{\uparrow}$&  3  & 66762 & 5942886 & 703977  &  747477    \\
& removing 50\%&  55.4$\pm$1.1 $\textcolor{red}{\uparrow}$& 6  &133524 &11885772  & 1407954   & 1494954    \\
&removing 70\%& 0$\pm 0$&8 &178032&- &  -  & -      \\
\hline
		\end{tabular}}
    \label{Main result1}
\end{table*}
}

%% file: figures/fig-1-ablation-performance-removal-rate.tex
\begin{figure}[t]
    \centering
    \includegraphics[width=\linewidth]{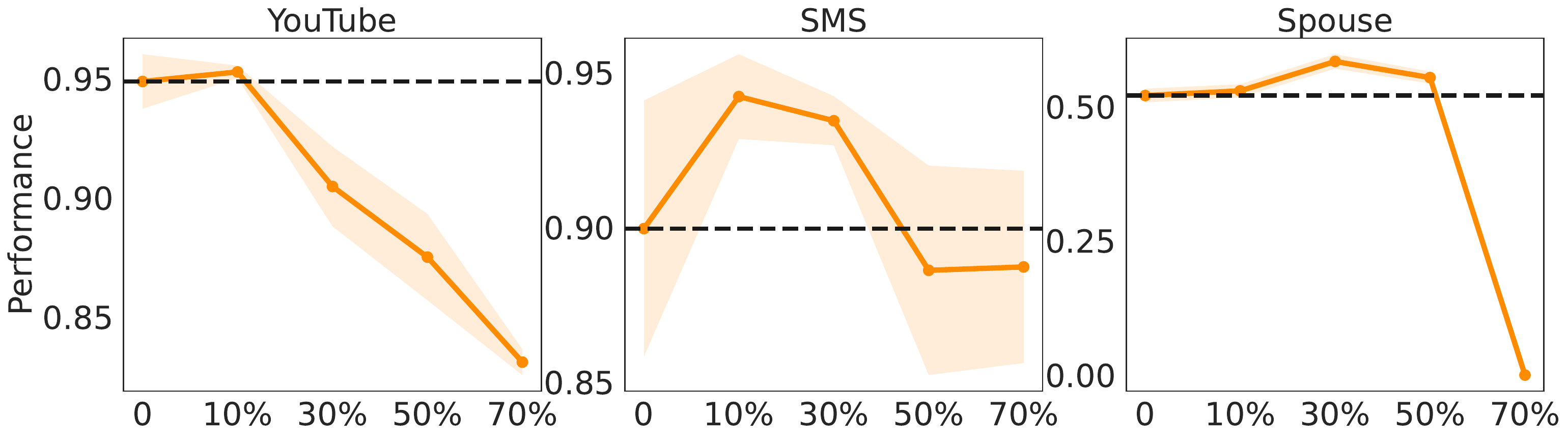}
    \caption{Performance on different removal rate. The dashed lines are the prompted weak supervisions without any removal. We plot the results of removing 10\%, 30\%, 50\%, 70\% of the labeling functions.}
    \label{pic3}
\end{figure}

%% file: sections/5-3-struct_gen.tex
\subsection{Effect of CosGen}

\input{figures/tables/tab-3-ablation-struct-comp}
Another core component of \methodname is Correlation Structure Generation (CosGen).
The main output of CosGen is the estimated structure $\mathcal{E}$, which contains all the labeling function pairs that are considered as correlated and potentially dependent on each other. 
In this subsection, we compare the CosGen module of our \methodname with weak supervision structure learning (WSSL) \citep{varma2019learning}, a state-of-the-art structure learning for programmatic weak supervision that learns a sparse structure from data.
We compare, evaluate and analyze both methods in terms of performance and efficiency. 
The performance comparisons are given in Table \ref{compared the structure part}. 

Table \ref{compared the structure part} shows that compared to PromptedWS without any consideration of correlations, passing the structure is extremely effective on Spouse dataset, improved 12.5 F1 score with CosGen and improved 11.8 F1 score using the WSSL. 
For SMS dataset, passing the structure also improves the results, boosting 3.4 Acc score and 4.5 Acc score using CosGen and the structure from WSSL respectively. 
Although both the structure from our refining module and the structure learned from data help to deal with dependencies and can effectively improve the performance, CosGen is much more efficient than WSSL.
The time and computational cost for CosGen is negligible while the time of learning structures using WSSL depends on the data and the number of LFs. 
\input{figures/tables/tab-4-ablation-runtime}
As shown in Table \ref{ta8}, structure learning with WSSL takes around 0.1s for YouTube and Spouse dataset and around 24s to learn a structure of SMS for WSSL. 
Note that running a label model without any structures for SMS dataset only takes less than 1s.
Compared to the label model run time, 24s for computing the structure alone should be considered as expensive and should not be ignored. 
The same holds for both YouTube and Spouse dataset: when using WSSL to learn a structure, the time spend on obtaining the structure could be longer than label model runtime.
In contrast, CosGen produces a structure that can be as effective as WSSL, but with negligible computation time.

%% file: figures/tables/tab-3-ablation-struct-comp.tex
\begin{table}[h]
    \centering
    \caption{Comparison of CosGen, weak supervision structure learning(WSSL) and without structures.}
    \resizebox{0.7\linewidth}{!}{
  		\begin{tabular}{|l|c|c |c |}
 \hline
&  \textbf{YouTube} & \textbf{SMS} & \textbf{Spouse} \\
 \hline
PromptedWS & \textbf{94.8}$\pm$1.2& 90.0$\pm$4.1&52.1$\pm$1.3\\
PromptedWS + WSSL& 93.0$\pm$1.3 &\textbf{94.5}$\pm$1.7 & 63.9$\pm$0.9\\
PromptedWS + CosGen& 94.2$\pm$0.5 &93.4$\pm$1.2 & \textbf{64.6}$\pm$0.5\\
 \hline
		\end{tabular}}
    \label{compared the structure part}
\end{table}

%% file: figures/tables/tab-4-ablation-runtime.tex
\begin{wraptable}{r}{6cm}
    \caption{WSSL runtime for inferring LFs structures. }
    \resizebox{\linewidth}{!}{
  		\begin{tabular}{|c|c|c|c|}
 \hline
& \textbf{YouTube} & \textbf{SMS} & \textbf{Spouse}\\
 \hline
time(s) & 0.11&24.76 &0.15 \\
 \hline
		\end{tabular}}
    \label{ta8}
\end{wraptable}

%% file: sections/5-4-augmentated_lfs.tex
\subsection{Handling high redundancy}
\label{sec:high_redundancy}
\input{figures/tables/tab-6-augmented}
To empirically evaluate the performance of \methodname in high redundancy scenarios, we conducted additional experiments using an original augmentation strategy for prompted LFs.
Using the same datasets, we translated the natural language prompts for each prompted LFs into a different language and then back into English. 
This process resulted in a new set of prompted LFs that are highly redundant with the original set.
We then merged these augmented prompted LFs with the original set, resulting in a larger and intentionally highly redundant set of prompted LFs.
The results of our experiments are shown in Table \ref{aug}, which demonstrates that \methodname achieved a significant improvement of 11.7 points over the PromptedWS baselines. 
This result indicates the effectiveness of \methodname in high redundancy scenarios.

%% file: figures/tables/tab-6-augmented.tex
{\renewcommand{\arraystretch}{1.2}
\begin{table}[h]
    \centering
    \caption{Comparisons with PromptedWS in augmented settings with redundent prompted LFs.}

  		\begin{tabular}{|l|c|c|c|}
 \hline
&{\splitcell{\textbf{YouTube(Acc)}}}&{\splitcell{\textbf{SMS(F1)}}}&{\splitcell{\textbf{Spouse(F1)}}}\\
\hline
\splitcell{PromptedWS\ \citep{smith2022language}}& 93.8$\pm$1.2& 89.0$\pm$3.6& 31.0$\pm$4.2\\
\splitcell{w/ \methodname}&\textbf{94.9}$\pm$1.1&\textbf{95.0}$\pm$1.0  &\textbf{58.9}$\pm$1.9 \\
 \hline
\end{tabular}
    \label{aug}
\end{table}
}
\input{figures/fig-2-runtime}

%% file: figures/fig-2-runtime.tex
\begin{figure*}[t]
    \centering
    \includegraphics[width=0.75\textwidth]{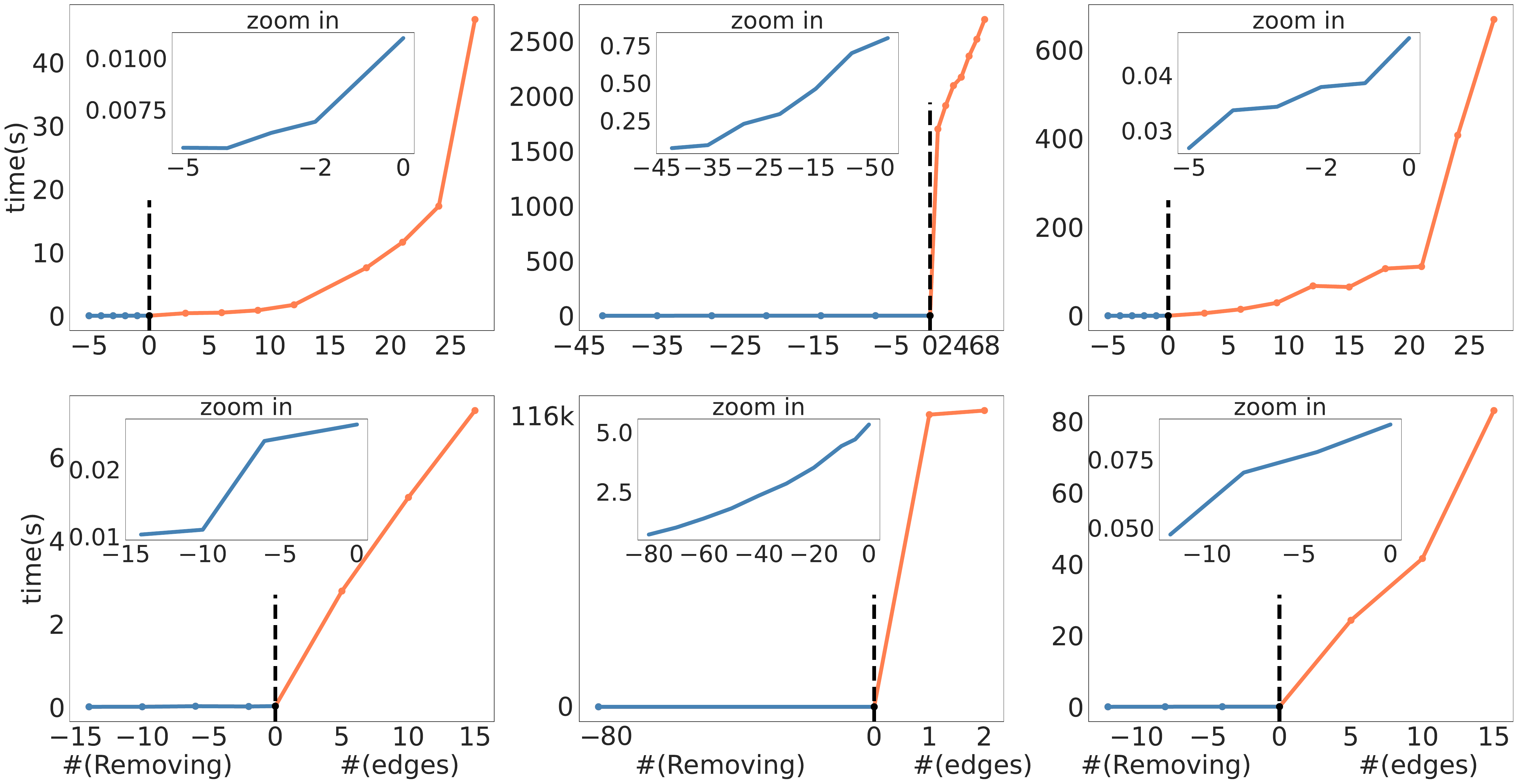}
    \caption{Label model running time for LaRe (left side of the dashed vertical line; plotted using blue) and CosGen (right side of dashed vertical line; plotted with orange). We zoom in the LaRe (blue plots) to better show the tendency in the subplots. From left to right columns are plots for Youtube, SMS and Spouse respectively. The first row describes experiments with original set of prompted LFs and the second row describes experiments with augmented LFs.}
    \label{pic5}
\end{figure*}

%% file: sections/5-5-effect_lm.tex
\subsection{Effect on label model efficiency}
\label{Efficiency}
In this subsection, we specifically exam how the computational time of the label model can change with \methodname.
In Figure \ref{pic5}, we plot the label model run time for both removing the LFs and feeding dependency structures into label model. 
A zero value on the horizontal axis represents PromptedWS without LaRe and CosGen, with negative values indicating the number of LFs removed and positive values indicating the number of edges.
 The vertical axis represents the run time of the label model.
We use blue color to represent number of LFs removed while and use orange color to represent number of edges in the structures. 
We observe that pass in the structure generally harms the efficiency of the label model.
Even passing a simple structure with just one edge can increase the computational time of the label model greatly.
For example, for the SMS dataset, passing a structure with only one edge leads to a surge in the running time from less than 1 second to around 1700 seconds, while for the SMS dataset with augmented LFs, it surges from 5 seconds to around $116 \times 10^3$ seconds. 
Similarly, as the number of edges increases, the computational time of the label model also increases, as seen in the YouTube and Spouse datasets.
 This tendency has been shown in both the YouTube and Spouse dataset, where when the number of edges increases, the running time of the label model tends to increase. 
 As for removing LFs, the running time saved on label model is not so remarkable unless the number of LFs are large. 
 Since for dataset with a few LFs, the label model is already fast enough even without removing any LFs. (Less than 1s for all the three dataset even with augmented LFs except for SMS dataset).
 In Figure \ref{pic5}, we zoom in on the running time for LF removal to see how running time changes when removing different number of LFs. 
Overall, our analysis suggests that both removing LFs and increasing sparsity in structures can improve the efficiency of the label model.
The choice should depend on the specific characteristics of the dataset and the trade-off between accuracy and efficiency.

%% file: sections/5-6-hyperparam_guide.tex
\subsection{Selecting hyperparameters}
In this subsection, we provide some intuition on how to select $m_r$ and $m_e$ for \methodname.
When selecting the hyperparameters for \methodname,  it is crucial to consider the balance between performance and computational costs. 
In scenarios with only a few LFs, occurrences of heavy redundancies among LFs also became rarer.
So removing too many labeling functions could harm performance.
In this case, the number to be removed $m_r$ should be relatively small. 
Since there are not many LFs to remove, there is not much to gain in computational time and resources. 
Here, performance should be the primary consideration.
When the total number of LFs is small,  a relatively sparser dependency structure among LFs may boost performance without significantly increasing run time. Thus, in this case, $m_e$ can be set to a non-zero value.

In contrast, if the number of LFs are large, there might be more substantial redundancies among LFs and thus removing some of them can improve the performance. 
In this case, We set $m_r$ to be relatively large. 
Moreover, since LFs removal saves computational time on the label model as well as computational cost for LLM inferences (such as costs on prompting), we can trade some performance for efficiency by selecting a large $m_r$.
We also prefer small $m_e$, the parameter controlling the number of edges of structure to relieve the computation cost.

%% file: sections/6-discussion.tex
\section{Conclusion and Discussion}
In this paper, we propose \methodname, a novel approach for efficient structure discovery and management in Prompted Weak Supervision.
Our approach asks large language models for information beyond the inference output, leveraging similarities in the embedding space to detect redundant LFs and learn the dependency structures among prompted LFs.
We demonstrated the effectiveness of our approach through extensive experiments and provide comprehensive ablation analysis for \methodname.
We believe that the \methodname is a valuable tool for weak supervision practitioners who wish to optimize their workflow for both efficiency and performance.
In future work, we plan to further investigate scalable and robust integration of LLMs into programmatic weak supervision workflows. \\

\noindent
\textbf{Ethical Considerations} One major concern for our method is unfair labeling due to spurious and sensitive feedback from LLMs.
As a result, supervised model using the data labeled by the proposed system may be subject to biased treatment in downstream applications.
Therefore it is essential to consider the potential biased output from LLMs and subject matter experts should also conduct careful moderation with the development dataset to better identify and contain the biases from LLMs.
Our proposed method aims to improve both the accuracy of probabilistic labeling and efficiency in structure estimation.
However, our methods may be unfairly used in applications that are subject to data privacy violations and the consequent labels or supervised-trained models.
It is crucial to apply careful scrutiny towards the detailed applications of our method.
While our method has the potentially to decrease the bias from LLMs, it is important to consider and carefully address these ethical challenges to ensure fair and responsible use. 

%% file: sections/8-acknowledgement.tex
\section*{Acknowledgements}
We would like to thank Jiayou Zhang for his helpful suggestions and comments.
This material is based on research sponsored by Defense Advanced Research
Projects Agency (DARPA) and Air Force Research Laboratory (AFRL) under agreement number
FA8750-19-2-1006. The U.S. Government is authorized to reproduce and distribute reprints for
Governmental purposes notwithstanding any copyright notation thereon. The views and conclusions
contained herein are those of the authors and should not be interpreted as necessarily representing
the official policies or endorsements, either expressed or implied, of Defense Advanced Research
Projects Agency (DARPA) and Air Force Research Laboratory (AFRL) or the U.S. Government. We gratefully acknowledge support from Google and Cisco.
Disclosure: Stephen Bach is an advisor to Snorkel AI, a company that provides software and services for data-centric artificial intelligence.

\noindent
\textbf{Authors’ Note.} The first two authors contributed equally. 
Co-first authors can prioritize their names when adding this paper’s reference to their resumes.